%% file: main.tex
\documentclass[10pt,twocolumn,letterpaper]{article}
\pdfoutput=1

\usepackage[pagenumbers]{wacv} 

\usepackage{graphicx}
\usepackage{amsmath}
\usepackage{amssymb}
\usepackage{booktabs}
\usepackage{multicol}
\usepackage{multirow}
\usepackage{xcolor,colortbl}
\usepackage{arydshln}
\usepackage{appendix}

\usepackage[pagebackref,breaklinks,colorlinks]{hyperref}

\usepackage[capitalize]{cleveref}
\crefname{section}{Sec.}{Secs.}
\Crefname{section}{Section}{Sections}
\Crefname{table}{Table}{Tables}
\crefname{table}{Tab.}{Tabs.}

\begin{document}

\title{
Enhanced Spatio-Temporal Context for Temporally Consistent Robust 3D Human Motion Recovery from Monocular Videos
}
\author{Sushovan Chanda\\
TCS Research\\
India\\
{\tt\small chanda.sushovan@tcs.com}
\and
Amogh Tiwari\\
CVIT, IIIT Hyderabad \\
India\\
{\tt\small amogh.tiwari@research.iiit.ac.in}
\and
Lokender Tiwari\\
TCS Research \\
India\\
{\tt\small lokender.tiwari@tcs.com}
\and
Brojeshwar Bhowmick\\
TCS Research \\
India\\
{\tt\small b.bhowmick@tcs.com}
\and
Avinash Sharma\\
CVIT, IIIT Hyderabad \\
India   \\
{\tt\small asharma@iiit.ac.in}
\and
Hrishav Barua\\
TCS Research \\
India\\
{\tt\small hrishav.smit5@gmail.com}
}

\twocolumn[{%
\renewcommand\twocolumn[1][]{#1}%
\maketitle
\begin{center}
    \centering
    \captionsetup{type=figure}
    \includegraphics[width=0.9\textwidth]{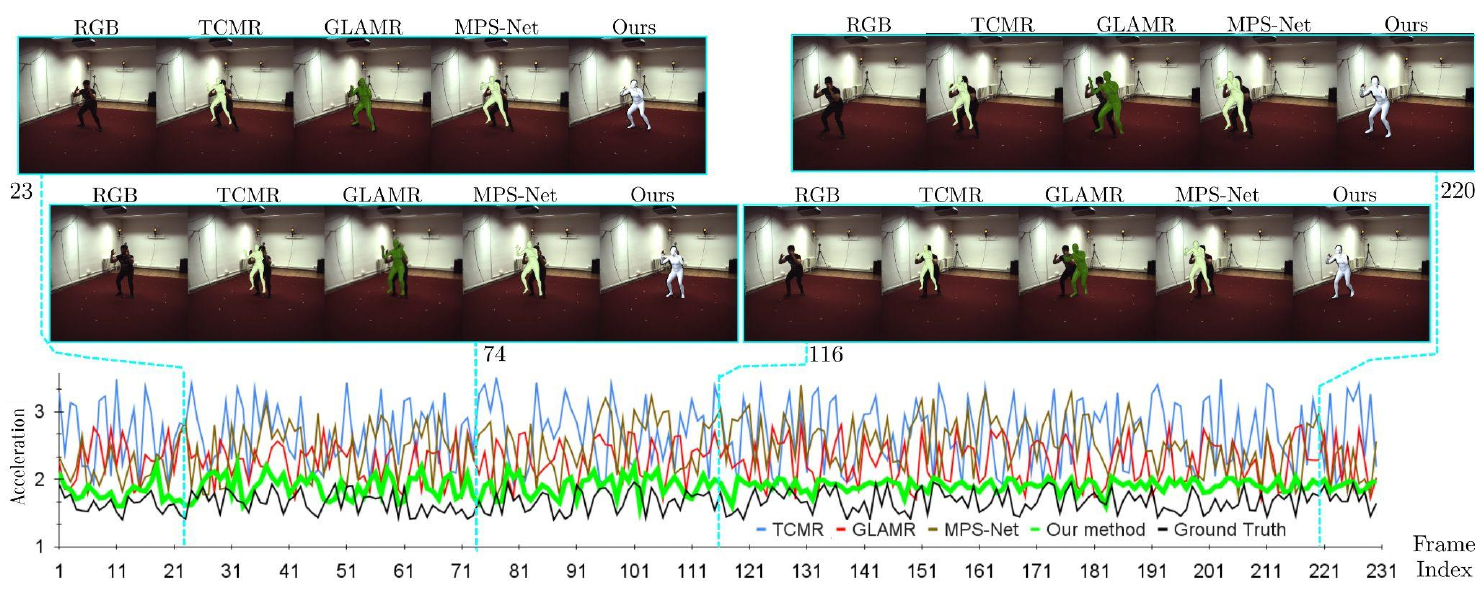}
    \caption{Our method yields superior and temporally consistent motion estimation. It can be observed that our method yield graph (green curve) significantly closer to the ground truth acceleration graph (black curve) compared to existing methods~\cite{choi2021beyond,yuan2022glamr,wei2022capturing} (result inferred on unseen test video from Human3.6M dataset~\cite{ionescu2013human3}).}
  \label{fig:teaser}
\end{center}%
}]

\begin{abstract}
Recovering temporally consistent 3D human body pose, shape and motion from a monocular video is a challenging task due to (self-)occlusions, poor lighting conditions, complex articulated body poses, depth ambiguity, and limited availability of annotated data. Further, doing a simple per-frame estimation is insufficient as it leads to jittery and implausible results. In this paper, we propose a novel method for temporally consistent motion estimation from a monocular video. Instead of using generic ResNet-like features, our method uses a body-aware feature representation and an independent per-frame pose and camera initialization over a temporal window followed by a novel spatio-temporal feature aggregation by using a combination of self-similarity and self-attention over the body-aware features and the per-frame initialization. Together, they yield enhanced spatio-temporal context for every frame by considering remaining past and future frames. These features are used to predict the pose and shape parameters of the human body model, which are further refined using an LSTM. Experimental results on the publicly available benchmark data show that our method attains significantly lower acceleration error and outperforms the existing state-of-the-art methods over all key quantitative evaluation metrics, including complex scenarios like partial occlusion, complex poses and even relatively low illumination. 
\end{abstract}

\input{sections/01_introduction}
\input{sections/02_related_works}
\input{sections/03_method}
\input{sections/04_exps}
\input{sections/05_discussion}
\input{sections/06_conclusion}

\onecolumn
\twocolumn
{\small
\bibliographystyle{ieee_fullname}
\bibliography{bibliography}
}


\end{document}

%% file: sections/01_introduction.tex
\section{Introduction}
\label{sec:intro}
Recovering 3D human body pose, shape and motion from a monocular video is an important task that has tremendous applications in augmented/virtual reality, healthcare, gaming, sports analysis, human-robot interaction in virtual environments, virtual try-on, etc. A lot of work has been done in estimating 3D body pose and shape from a single-image \cite{georgakis2020hierarchical,kanazawa2018end,kolotouros2019learning,omran2018neural,pavlakos2018learning} by learning to regress the explicit 3D skeleton or parametric 3D body model (e.g., SMPL~\cite{loper2015smpl}). However, many applications such as human motion analysis, sports analytics, behavior analysis, etc., critically depend on the temporal consistency of human motion where single-image-based methods seem to fail frequently. Temporally consistent 3D human pose, shape and motion estimation from a monocular video is a challenging task due to (self-) occlusions, poor lighting conditions, complex articulated body poses, depth ambiguity, and limited availability of annotated data. 
Efforts on monocular video-based motion estimation~\cite{kocabas2020vibe,luo20203d,choi2021beyond, lee2021uncertainty, pavllo:videopose3d:2019, tripathi2020posenet3d} typically introduce a CNN or RNN module to perform spatio-temporal feature aggregation from neighboring frames followed by SMPL \cite{loper2015smpl} parameters regression, thus modeling relatively local temporal coherence.  However, these methods tend to fail while capturing long-term temporal dynamics and show poor performance when the body is under partial occlusion. Some of the recent works~\cite{rempe2021humor,henemf,marsot2021structured,yuan2022glamr} also attempt to model the generative space of motion modeling using Conditional VAEs, often followed by a global, non-learning-based optimization at inference time using the entire video. Such global optimization is also used in a very recent work in~\cite{zeng2022smoothnet} with a plug-and-play post-processing step for improving the existing methods by exploiting long-term temporal dependencies for human motion estimation. However, due to the post-processing over the entire sequence, such methods find limited applicability to real-world scenarios.

A highly relevant recent work, MPS-Net~\cite{wei2022capturing}, attempts to attain a good balance between local to global temporal coherence using their MOtion Continuity Attention (MOCA) module. More specifically, their method explicitly models the visual feature similarity across RGB frames and uses it to guide the learning of the self-attention module for spatio-temporal feature learning. MOCA enables focusing on an adaptive neighborhood range for identifying the motion continuity dependencies. This is followed by a Hierarchical Attentive Feature Integration (HAFI) module to achieve local to global temporal feature aggregation through which they achieve SOTA performance. Nevertheless, similar to the majority of the existing methods, they use ResNet~\cite{he2016deep}-like generic deep features extracted from the RGB frames. However, such generic feature representations do not exploit the prior knowledge of human appearance(that the 3D human body has a fixed topology and can be represented by a parametric model). Additionally,~\cite{wei2022capturing} do not exploit per-frame pose and shape initialization and uses a computationally heavy Hierarchical Attentive Feature Integration (HAFI) module. 
Finally, they only perform per-frame prediction using the aggregated spatio-temporal features, thereby completely neglecting the joint estimation performed by existing methods.

In this paper, we propose a holistic method that exploits enhanced spatio-temporal context and recovers temporally consistent 3D human pose/shape from monocular video. At first, we select a set of continuous frames in a temporal window and pass it to the {\it Initialization module} which extracts the body-aware deep features from individual frames and in-parallel predict initial per-frame estimates of body pose/shape and camera pose using an off-the-shelf method. Subsequently, we pass these initial estimates and features to novel {\it Spatio-Temporal feature Aggregation (STA) module} for recovering enhanced spatio-temporal features.  Finally, we employ our novel {\it Motion estimation and Refinement module} to obtain temporally consistent pose/shape estimation using these enhanced features.  \autoref{fig:method_fig} provides outline of our method. 

In regard to functionality/relevance of these modules, the initialization module extracts a body-aware feature representation~\cite{neverova2020continuous} for each frame of the local non-overlapping temporal frame window, instead of the generic ResNet feature used by existing methods and the independent per-frame pose and camera initialization estimated using~\cite{goel2023humans}. 
This provides a strong spatial prior to our method.  Further, our proposed novel STA module computes the self-similarity and the self-attention on initial spatial priors provided by the previous module. In particular, the self-similarity between the body-aware features in a temporal window helps us to correlate the body parts across frames even in the presence of occlusion. Similarly, the self-similarity among the pose parameters and the cameras reveals the continuity of the human motion along with the camera consistency.  We also use self-attention on the camera parameters and the body-aware features.
Together, they yield spatio-temporal aggregated features for every frame by considering the remaining past and future frames inside the window.  
Here, the joint characteristics of the self-similarity and the attention map find the more appropriate range in the input video to reveal the long-horizon context. 
Finally, our novel motion estimation and refinement module first predicts the per-frame coarse estimation of pose/shape using the spatio-temporally aggregated features from the STA module and subsequently passes it to an LSTM-based joint-temporal refinement network to recover the temporally consistent robust prediction of pose/shape estimates. In order to generate continuous predictions near the temporal window boundaries, we average the pose/shape parameters for consecutive border frames across neighboring windows. 
We empirically observed that applying LSTM-based joint refinement on pose/shape yields superior performance instead of applying it on STA features and then predicting pose/shape parameters (see~\autoref{sec:ablation}). 

As a cumulative effect, our method produces significantly lower acceleration errors in comparison to SOTA methods (see~\autoref{sec:quantres}).~\autoref{fig:teaser} shows a plot of acceleration where our method yields the acceleration curve (in green) closest to the ground truth acceleration curve (in black).  
Moreover, owing to our enhanced spatio-temporal context and motion refinement, our method significantly outperforms the state-of-the-art (SOTA) methods even in relatively poor illumination and severe occlusion (please refer to~\autoref{sec:qualres}). 

\definecolor{aqua}{rgb}{0.0, 1.0, 1.0}
\definecolor{aquamarine}{rgb}{0.5, 1.0, 0.83}
\definecolor{ao}{rgb}{0.0, 0.0, 1.0}
\definecolor{blue}{rgb}{0.0, 0.0, 1.0}
\definecolor{blue(ryb)}{rgb}{0.01, 0.28, 1.0}
\definecolor{bisque}{rgb}{1.0, 0.89, 0.77} 	
\definecolor{bananamania}{rgb}{0.98, 0.91, 0.71}
\definecolor{bananayellow}{rgb}{1.0, 0.88, 0.21}
\definecolor{arylideyellow}{rgb}{0.91, 0.84, 0.42}
\definecolor{apricot}{rgb}{0.98, 0.81, 0.69}
\definecolor{aureolin}{rgb}{0.99, 0.93, 0.0}
\definecolor{awesome}{rgb}{1.0, 0.13, 0.32}
\definecolor{bittersweet}{rgb}{1.0, 0.44, 0.37}
\definecolor{dandelion}{rgb}{0.94, 0.88, 0.19}
\definecolor{buff}{rgb}{0.94, 0.86, 0.51}
\definecolor{cyan(process)}{rgb}{0.0, 0.72, 0.92}
\definecolor{cottoncandy}{rgb}{1.0, 0.74, 0.85}
\definecolor{corn}{rgb}{0.98, 0.93, 0.36}
\definecolor{flavescent}{rgb}{0.97, 0.91, 0.56}
\definecolor{babyblue}{rgb}{0.54, 0.81, 0.94}
\definecolor{beaublue}{rgb}{0.74, 0.83, 0.9}
\definecolor{blizzardblue}{rgb}{0.67, 0.9, 0.93}
\definecolor{columbiablue}{rgb}{0.61, 0.87, 1.0}

%% file: sections/02_related_works.tex
\section{Related Works}
\textbf{Image based 3D human pose, shape and motion estimation}: 
Existing methods either solve for the parameters of SMPL \cite{loper2015smpl} from the images or directly regress the coordinates of a 3D human mesh \cite{Gyeongsik_nonparametric}. \cite{kanazawa2018end, kolotouros2019learning, omran2018neural} are some of the early succesful works for human pose and shape estimation from monocular images. 

HyBrik \cite{li2021hybrik} and KAMA \cite{iqbal2021kama} leverage  3D key points for the 3D mesh reconstruction.  In particular, HyBrik uses twist and swing decomposition for transforming the 3D joints to relative body-part rotations. Instead of full body, methods like HoloPose \cite{guler2019holopose} and PARE \cite{kocabas2021pare} have introduced parts parts-based model. While HoloPose does part-based parameter regression, PARE uses a part-guided attention mechanism for exploiting the visibility of individual body parts and predicting the occluded parts using neighboring body-part information. While these methods are quite effective for estimating the 3D pose and shape from images, they are not capable of producing temporally consistent 3D human motion from video by frame-based processing. \\

\textbf{Video based 3D human and pose estimation}:
Recently, a considerable amount of work has been carried out to address the challenge of temporally consistent 3D human pose and shape estimation from video. For instance, HMMR \cite{kanazawa2019learning} trains a temporal encoder that learns a representation of 3D human dynamics from a temporal context of image features. Along with 3D human pose and shape, 
such representation is also used for capturing the changes in the pose in the nearby past and future frames. Similarly, VIBE \cite{kocabas2020vibe} proposes a temporal encoder that encodes static features into a series of temporally correlated latent features and feeds them to a regressor to estimate the SMPL parameters. MEVA \cite{luo20203d} uses a two-stage model that first captures the coarse overall 3D human motion followed by a residual estimation that adds back person-specific motion details. However, these methods fail to reconstruct the humans under partial occlusions. In a recent work by Choi \textit{et al.} \cite{choi2021beyond}, GRU-based temporal encoders are used with different encoding strategies to learn better temporal features from images. Also, they propose a feature integration from the three encoders for the SMPL parameter regressor. GRU-based techniques can only deal with local neighborhoods which makes it difficult for them to learn long-range dependencies. Hence, ~\cite{baradel2022posebert, motionbert2022} uses a transformer to learn long-range temporal dependencies. However, such methods require a large number of consecutive frames (around 250), making them slower. Another class of methods like HuMoR \cite{rempe2021humor} and GLAMR \cite{yuan2022glamr} use variational autoencoder which takes single frame-based human pose estimates to predict the human motion sequence in an auto-regressive way followed by a non-learning based global optimization on the human pose and trajectory obtained from the entire video for temporal refinement. Similarly, SmoothNet \cite{zeng2022smoothnet} also does a global optimization on the estimated trajectory of any human pose estimation method to improve their temporal continuity. The global optimization in the test time limits the applicability of such methods.  In a recent attempt, MPS-Net \cite{wei2022capturing} tries to produce locally global temporal coherence using a motion continuity attention module (see~\autoref{sec:intro} for more details on MPS-Net~\cite{wei2022capturing}).

%% file: sections/03_method.tex
\section{Method}

\begin{figure*}
    \centering
    \includegraphics[width=\textwidth]{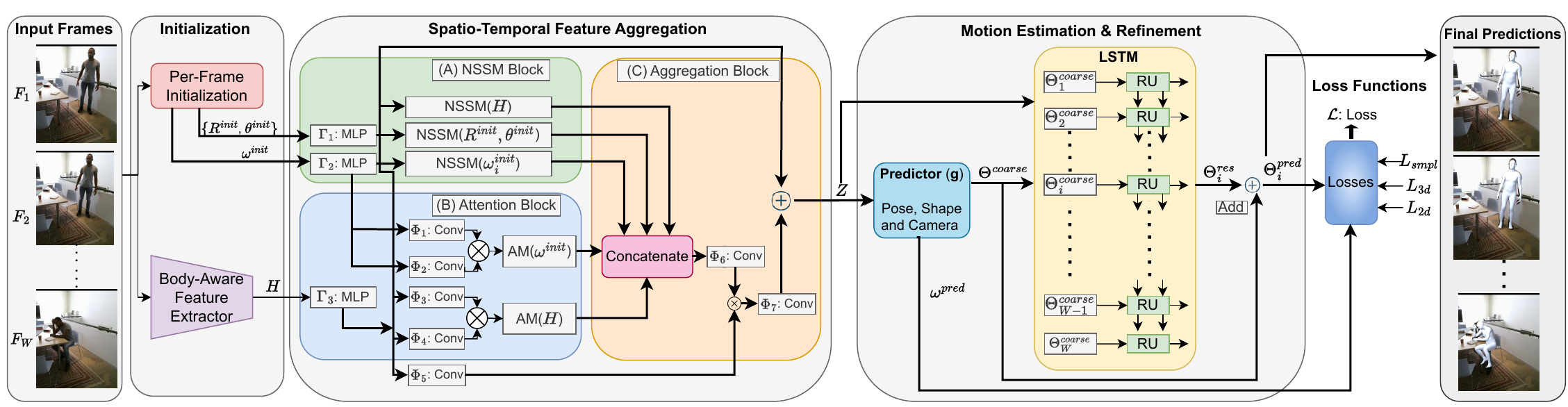}
    \caption{Overview of our proposed method.}
    \label{fig:method_fig}
\end{figure*}

\begin{figure}
    \centering
    \includegraphics[width=\linewidth]{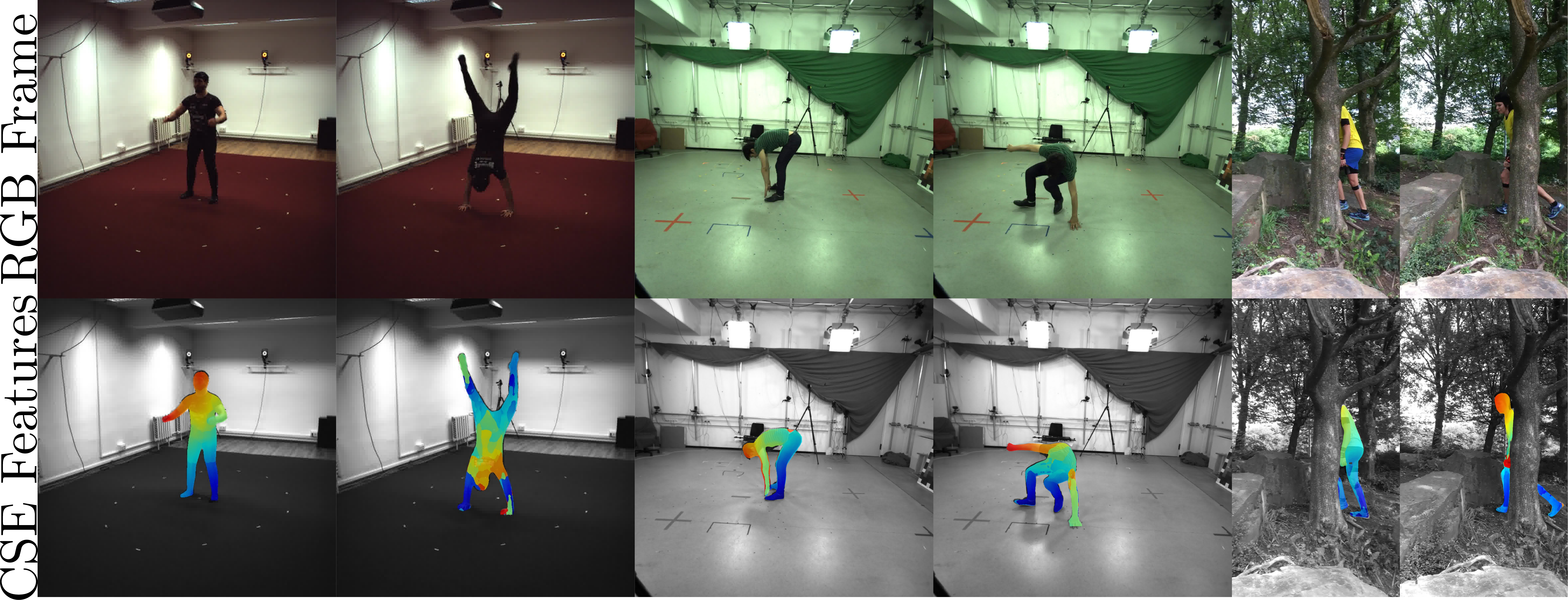}
    \caption{Sample three-channel visualization of CSE Embeddings.}
    \label{fig:cse_fig}
\end{figure}

In this section, we provide a detailed overview of the key modules of our proposed method. As discussed in~\autoref{sec:intro} (and outlined in~\autoref{fig:method_fig}), our method takes a set of consecutive frames as input and feeds it to the three key modules, namely, initialization, spatio-temporal feature aggregation and motion prediction \& refinement, to predict temporally consistent body pose and shape parameters of SMPL~\cite{loper2015smpl}, a statistical body model.

More specifically, given an input video $V=\{F_i\}_{i=1}^N$ composed of N frames, with $F_i$ representing the $i^{th}$ frame, we aim to recover SMPL-based human body pose and shape parameters for each frame, i.e., $\Theta^{pred}_i = \{T_i, R_ic, \theta_i, \beta_i\}$. Here, $T_i \in \mathbb{R}^{3}$ and $R_i \in \mathbb{R}^{3}$ represents the translation and rotation (in axis-angle format) of the root joint, $\theta_i \in \mathbb{R}^{23\times3}$ represents the relative rotations of the remaining 23 joints while $\beta_i \in \mathbb{R}^{10}$ represents body shape parameters. Please note that we sample a temporal window (a subset of continuous frames) of size $W$ (we choose $W=16$) from the input video and learn/infer over it instead of doing inference on all frames in the video sequence. 
\subsection{Initialization}
\noindent {\bf Per-frame Body Pose and Camera Estimation:} We perform independent estimation of per-frame body pose ($\theta^{init}$) and camera ($\omega^{init}$) parameters using a SOTA method (HMR2.0~\cite{goel2023humans}) and feed it as initialization to our STA module.   \\
~\\
\noindent {\bf Body-aware Spatial Feature Extraction:}
Recently, Continuous Surface Embedding (CSE)~\cite{neverova2020continuous} was proposed to learn body-aware feature representation for obtaining dense correspondences across images of humans. CSE predicts, per pixel $16$-dimensional embedding vector (associated with the corresponding vertex in the parametric human mesh), thereby establishing dense correspondences between image pixels and 3D mesh surface, even in the presence of severe illumination conditions and (self-) occlusions. ~\autoref{fig:cse_fig} shows the color-coded visualization of CSE embeddings demonstrating its robustness to severe illumination and/or occlusion scenarios. Thus, we propose to extract and use the $16$-dimensional body-aware spatial features $H=\{H_i\}_{i=1}^N$ using a pre-trained CSE encoder for each frame $F_i$, such that:

\begin{equation}
    H_i = \Psi(F_i)
\end{equation}
where, $H_i \in \mathbb{R}^{112\times 112 \times16}$. 

\subsection{Spatio-Temporal Feature Aggregation (STA)}
The spatial features $H_i$ extracted from each frame can be directly regressed to estimate per-frame motion and shape parameters. However, this typically leads to jittery and implausible motion estimates as the predictions are not temporally consistent. 
One possible remedy to this is to use self-attention across frames in a temporal window~\cite{Wang2017NonlocalNN}. 
Interestingly, \cite{wei2022capturing} showed that a regular attention network is unreliable and can give high attention scores between temporally distant frames which would lead to inaccurate results. They address this problem by using a \textbf{Normalized Self-Similarity Matrix (NSSM)} in their MOCA module. Nevertheless, their method only exploited the spatial features for such self-attention guidance. Instead, as per the recent trend of exploiting per-frame pose initialization~\cite{yuan2022glamr, li2022dnd}, we propose to encode additional information to our temporal features in terms of initial estimates of body pose and camera parameters. More specifically, we obtain for each $i^{th}$ frame the initial pose/shape and camera parameters using~\cite{goel2023humans} as: $\Theta_{i}^{init} = \{T_i, R_i, \theta_i, \beta_i\}$ and camera parameters $\omega^{init}_i \in \mathbb{R}^3$ (assume a weak perspective camera model). 
It is important to note that we represent rotation using the 6-dimensional vector representation~\cite{zhou2019continuity} and then flatten them into a single 144-dimensional vector to recover body pose as: $[R_i,\theta_i] \in \mathbb{R}^ {144}$

Our STA module has three key blocks: (1) Frame-wise Similarity Computation, (2) Frame-wise self-attention, and (3) Feature Aggregation. \\
The first block deals with the computation of the three $\{W \times W\}$ self-similarity matrices, namely, NSSM ($H$) for the Body-aware spatial features, NSSM ($[R, \theta]$) for initial body pose estimates and NSSM ($\omega^{init}$) for initial camera estimates. More specifically, we uplift $[R_i, \theta_i]$ and $\omega^{init}_i \in R^3$ to 512 dimensions using linear layers $\Gamma_1$ and $\Gamma_2$ and similarly transform the spatial feature $H_i$ to 2048 dimensions using $\Gamma_3$. 
These multiple NSSMs help us to correlate the frames based upon body parts appearance, body pose, and cameras thereby giving robustness to occlusions as well as revealing the continuity of the human motion along with the camera consistency.

\noindent The second block obtains a self-attention map on our spatial features i.e., AM ($H$) and initial camera estimates i.e., AM ($\omega^{init}$), respectively. When applying self-attention on our spatial features $H_i \in \mathbb{R}^{112\times112\times16}$, first we transform them to 2048 dimension using a linear layer $\Gamma_3$ , and later down-sample them to $\mathbb{R}^{N\times 1024}$ by learning two different $1 \times 1$ convolution layers $\Phi_3$ and $\Phi_4$. Similarly, when applying self-attention on the initial camera estimates, we first uplift this vector to 512 dimension vector using an MLP $\Gamma_2$ and subsequently learn two different $1 \times 1$ convolution layers $\Phi_1$ and $\Phi_2$. This self-attention on the camera parameters and the body-aware features help us adaptively find the range which is important to capture the temporal smoothness.

Finally, the feature aggregation block first concatenates all the attention and NSSM maps to get a $W \times W \times 5$ tensor and later resize it to $W\times W$ matrix using a $1\times 1$ convolution layer ($\Phi_6$). This $W\times W$ represents the consolidated similarity between frames across the window. This feature is subsequently multiplied with the down-sampled spatial features (of 1024 dimension obtained by $\Phi_5$) and the result is then uplifted (using convolution layer $\Phi_7$) to get $Y \in \mathbb{R}^{W\times 2048}$. Thus, together, they yield spatio-temporal aggregated features for every frame by considering the remaining past and future frames inside the window. The per-frame temporally aggregated feature $Y_i$ is finally added to the spatial features to get the spatio-temporally aggregated features $Z_i$ for $i^{th}$ frame as:
\begin{equation}
    Z_i = H_i + Y_i.
\end{equation}
\subsection{ Motion Estimation \&  Refinement}
Once we have the spatio-temporal features $Z_i$, we obtain an independent coarse pose/shape, and camera estimation for each frame using predictor network ($g$)
\begin{equation}
    \Theta^{coarse}_i, \omega^{pred}_i = g(Z_i)
\end{equation}
where $g$ predicts the SMPL parameters i.e., $\Theta^{coarse}_i \in \mathbb{R}^{85}$ and the camera parameters i.e., $\omega^{pred}_i \in \mathbb{R}^3$ for frame $F_i$. 

We propose to further refine these estimated independent coarse poses and shapes (obtained using spatiotemporally aggregated features) using an LSTM \cite{sepp1997lstm} based joint \textit{residual} prediction. The LSTM $\zeta$ takes as input the features $Z_i$ and coasre SMPL pose estimates $\Theta^{coarse}_i$ and predicts the residual $\Theta^{res}_i \in \mathbb{R}^{85}$ , which is subsequently added to $\Theta^{coarse}_i$ in order to recover the refined pose and shape parameters $\Theta^{pred}$.
\begin{eqnarray}
\Theta^{res}_i = \zeta(Z_i, \Theta_{i}^{coarse}) \\
\Theta^{pred}_i = \Theta^{coarse}_i + \Theta^{res}_i
\end{eqnarray}
To ensure temporally consistent predictions at the window boundaries, we average the pose and shape parameter estimates of bordering frames across the neighboring windows.
\subsection{Loss Functions}

Similar to existing literature~\cite{wei2022capturing, kanazawa2018end, kocabas2020vibe}, we adopt loss functions on body pose and shape ($L_{SMPL}$), 3D joint coordinates ($L_{3D}$), and 2D joint coordinates ($L_{2D}$) obtained with predicted weak-perspective camera parameters ($\omega^{pose}$). These loss functions are briefly explained below.  
\begin{equation}
    \begin{aligned}
    L_{SMPL} & = \lambda_{shape} ||\hat{\beta}_i - \beta_i||_2 \\
    & + \lambda_{pose} ||\{\hat{R_i}, \hat{\theta_i}\} - \{R_i, \theta_i\} ||_2
    \end{aligned}
    \label{eq:smpl_loss}
\end{equation}
where $\beta_i$ and $\{R_i, \theta_i\}$ respectively are the predicted pose and shape parameters for the $i^{th}$ frame, and $\hat{\beta_i}$, $\{\hat{R_i}, \hat{\theta}_i\}$ are the corresponding ground-truths.

\begin{equation}
    L_{3D} = ||\hat{J^c_i} - J^c_i||_2
    \label{eq:3d_joint_loss}
\end{equation}
where $J^c_i$ represents predicted the 3D joint coordinates for the $i^{th}$ frame and $\hat{J^c_i}$ are the corresponding ground-truth 3D joint coordinates.

\begin{equation}
    L_{2D} = ||\hat{x_i} - \Pi(J^c_i)||_2
    \label{eq:2d_joint_loss}
\end{equation}
where $\hat{x}_i$ represents the ground-truth 2D keypoints for the $i^{th}$ frame and $\Pi$ represents the 3D-2D projection obtained from the predicted camera parameters $\omega^{pred}$.

The final loss function is a  linear combination of these losses defined as:
\begin{equation}
    L_{final} = \lambda_1L_{SMPL} + \lambda_2L_{3D} + \lambda_3L_{2D} 
    \label{eq:final_loss}
\end{equation}

It is important to note that our model is trained in an end-to-end trainable fashion where $L_{final}$ is applied on the final predicted pose and shape parameters obtained from LSTM $\zeta$. There is no separate training performed for the coarse estimation predictor $g$. 

%% file: sections/04_exps.tex

\section{Experiments and Results}
\label{sec:experiments}
\input{tables/tab_quant}

\noindent \textbf{Datasets Details:} We adopt the same train/test splits of Human3.6M~\cite{ionescu2013human3}, 3DPW~\cite{von2018recovering} and MPI-INF-3DHP~\cite{mehta2017monocular} datasets used by existing work~\cite{kocabas2020vibe, choi2021beyond, wei2022capturing}. \textit{Human3.6M} is a large scale dataset containing video sequences with corresponding 3D pose annotations of various subjects performing different actions like discussion, smoking, talking on the phone etc. Similar to existing work ~\cite{kocabas2020vibe, choi2021beyond, wei2022capturing}, we use the sub-sampled dataset (25 FPS) for our experiments. \textit{MPI-INF-3DHP} contains 8 subjects with 16 videos per subject. It is captured in a combination of indoor and outdoor settings with actions ranging from walking and sitting, to complex dynamic actions like exercising. It is captured by a markerless motion capture system using a multi-view camera setup. \textit{3DPW} is an in-the-wild dataset, captured with a moving cell-phone camera. It uses inertial measurement unit (IMU) sensors patched to the human body parts to calculate the ground-truth SMPL~\cite{loper2015smpl} parameters. It contains 60 video sequences with 18 3D models in different clothing, performing daily-life activities like walking, buying vegetables etc. Further, in order to evaluate the generalization ability of our method to unseen data, we use three additional datasets: Fitness-AQA~\cite{parmar2022domain}, PROX~\cite{hassan2019resolving} and i3DB~\cite{monszpart2019imapper}. These datasets contain sequences having actions/motion fairly different from our training datasets. Fitness-AQA contains videos of subjects lifting weights in a gym, which leads to self-occlusion and complex body poses, while PROX and i3DB contain video sequences of humans interacting with objects in an indoor setting like a room/office.

\noindent\textbf{Evaluation Metrics:} We use the standard evaluation metrics used in existing literature~\cite{kocabas2020vibe, luo20203d, choi2021beyond, wei2022capturing} to evaluate our method's performance. Specifically, we use the mean per joint position error (MPJPE), Procrustes-aligned mean per joint position error (PA-MPJPE), mean per vertex position error (MPVPE) and acceleration error (ACC-ERR). MPJPE is defined as the mean of the Euclidean distances between the ground truth and the predicted joint positions. PA-MPJPE is defined as the MPJPE computed after using Procrustes alignment (PA) to solve for translation, scale and rotation between the estimated body and the ground truth. MPVPE is given by the mean of the Euclidean distances between the ground truth and the predicted vertex positions of each vertex in the SMPL~\cite{loper2015smpl} body model constructed using the predicted pose/shape parameters. Finally, (ACC-ERR) is defined as the mean difference between the accelerations of the ground truth and predicted 3D joints. Specifically, the change in position of the 3D joints in unit time (i.e. across two consecutive frames) gives us the velocity of the joints, and the change in velocity in unit time gives us the acceleration. Acceleration error is then measured by finding the difference between the groundtruth and predicted accelerations. MPJPE, PA-MPJPE and MPVPE are measured in millimeters ($mm$) and express the fidelity of the estimated 3D pose and shape. While ACC-ERR, measured in $mm/t^{2}$ (where $t$ denotes unit time - the time interval between two consecutive frames) expresses the temporal consistency of the estimation.

\input{tables/tab_generalization}
~\\
\noindent\textbf{Implementation Details:}
We obtain the body aware features and per-frame pose/camera initializations using the pre-trained CSE~\cite{neverova2020continuous} and HMR2.0~\cite{goel2023humans} models, respectively. Similar to existing work~\cite{kocabas2020vibe, choi2021beyond, wei2022capturing}, we initialize our pose, shape, and camera predictor in the motion estimation and refinement module with the pre-trained SPIN~\cite{kolotouros2019learning} checkpoint. In the same module, the LSTM has 3 layers and uses $2048$ as the hidden feature size. Training is performed for 35 epochs with a mini-batch size of 32 and an initial learning rate of $5 \times 10^{-5}$. The learning rate is reduced by a factor of 10 every time the 3D pose accuracy does not improve for the 5 consecutive epochs. Adam Solver~\cite{kingma2014adam} is used for optimization. 
 For our experiments, we use a window size of 16 (see ~\autoref{tab:ablation_window_size} for discussion on choice of window size). We set the coefficients for $L_{SMPL}$, $L_{3D}$ and $L_{2D}$ to $300.0$, $0.06$ and $60.0$ respectively. Training is done for 35 epochs and takes about 7 hours using 3 NVIDIA RTX A-6000 GPUs.

\subsection{Quantitative Results}
\label{sec:quantres}
\begin{figure*}[h!]
    \centering
\includegraphics[width=\textwidth]{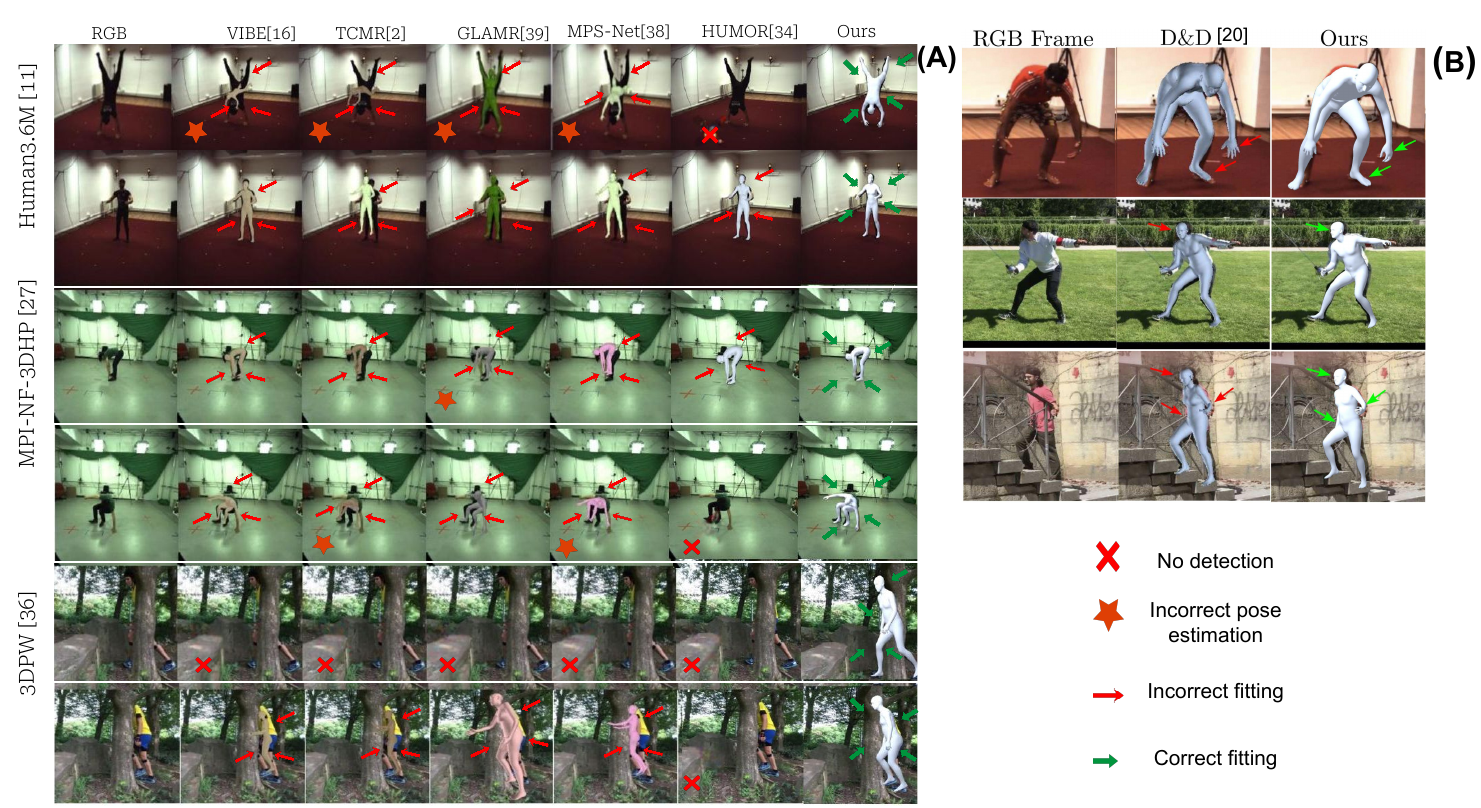}

\caption{(A) We show the estimated pose overlaid on the frames of the sample videos from Human3.6M~\cite{ionescu2013human3}, 3DPW~\cite{von2018recovering}, and MPI-INF-3DHP datasets~\cite{mehta2017monocular}. (B) Similar results are shown in comparison to D\&D~\cite{li2022dnd}.}
    \label{fig:qltv01}
\end{figure*}
\begin{figure*}
    \centering
    \includegraphics[width=\textwidth]{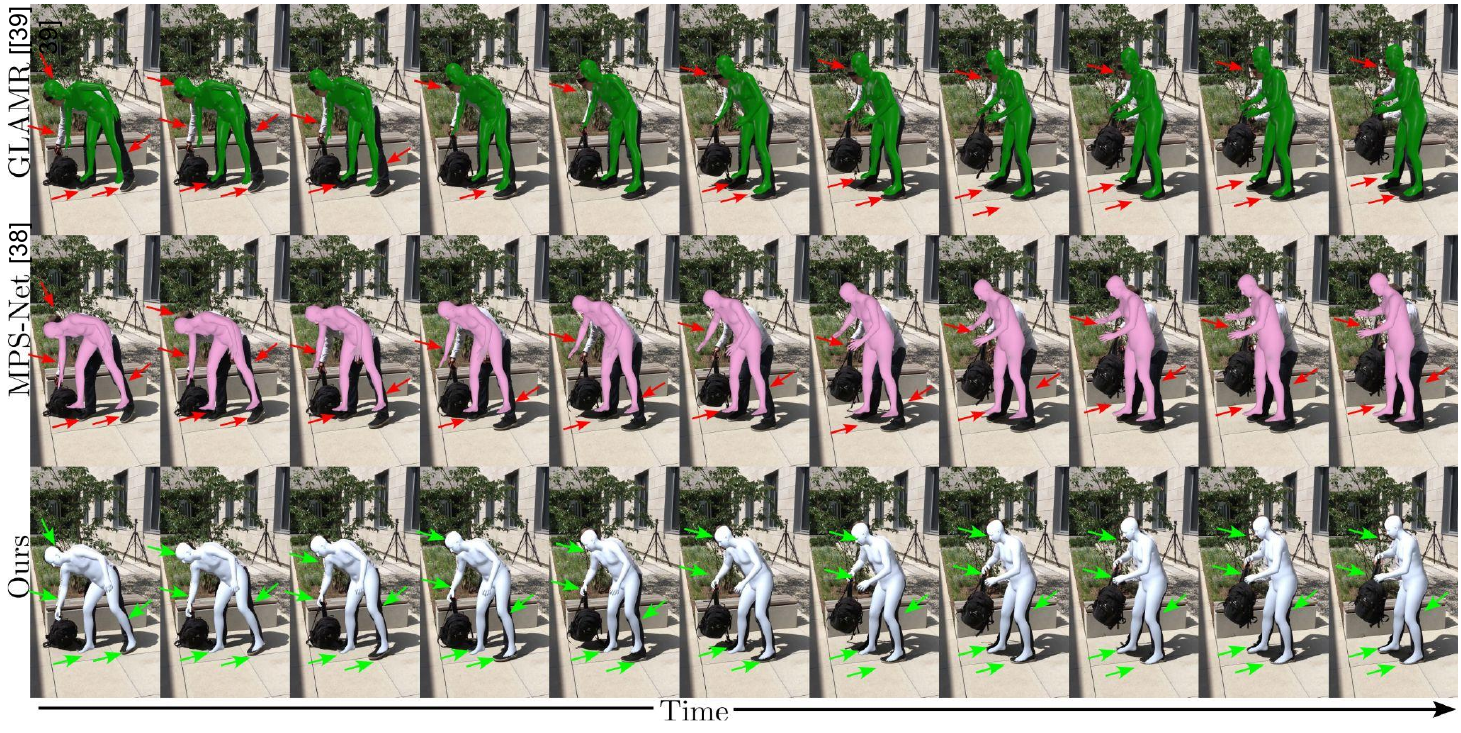}
    \caption{Qualitative comparison across frames on a sequence of 3DPW dataset. The green arrows show the improved regions compared to the red ones. Our method is able to achieve more accurate and more temporally consistent SMPL fitting.
}
    \label{fig:qltv02_temp_const}
\end{figure*}
\begin{figure*}[h!]
    \centering
    \includegraphics[width=\textwidth]{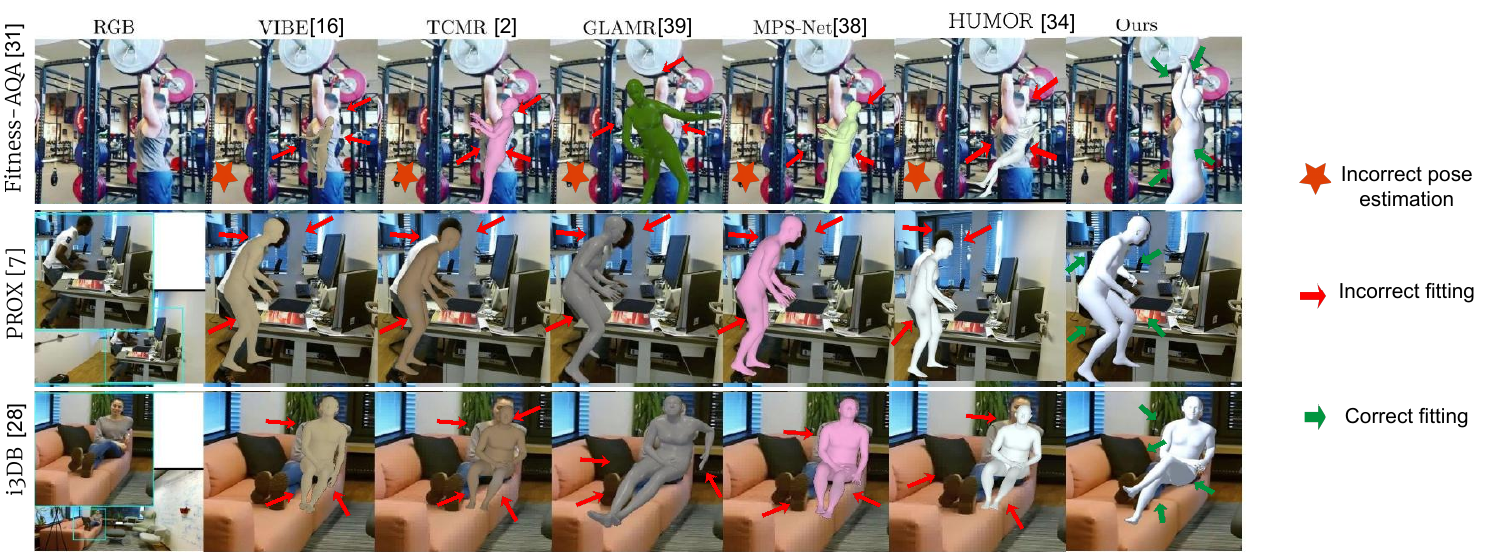}
    \caption{Qualitative comparison on unseen datasets.}
    \label{fig:qltv03_ood}
\end{figure*}
%
\input{tables/tab_ablation_new}

\input{tables/tab_ablation_diff_init}
%
\input{tables/tab_ablation_window_size}
%
~\autoref{tab:quant} provides a quantitative comparison between our method and existing SOTA monocular video-based methods. All methods use the same 3D datasets as ours with standard training/test split, except GLAMR~\cite{yuan2022glamr} (which uses Human3.6M, 3DPW and AMASS datasets) and D\&D~\cite{li2022dnd} (which trains individually on Human3.6M and 3DPW). Some of these methods also utilize additional 2D datasets for training and we used their pre-trained model for comparison. However, we only rely on the 3D datasets for training. It can be observed from~\autoref{tab:quant} that our method significantly outperforms existing methods over all metrics across datasets, demonstrating the superiority of our method.
\subsection{Qualitative Results}
\label{sec:qualres}

\autoref{fig:qltv01} visualize qualitative comparison with monocular video-based SOTA methods. More specifically, row-1 shows a complex pose under poor illumination. It can be observed that while all other methods fail to recover the body pose, our method successfully recovers the body pose reliably. In rows 2-4, it can be observed that while other methods are also able to recover the pose, our method provides more accurate SMPL fitting. In the last two rows, we demonstrate results on even more challenging cases involving significant occlusion, where most existing methods fail even to detect the human. However, our method not only detects the human but also provides a reasonable SMPL fitting.
Additionally, in~\autoref{fig:qltv02_temp_const}, we demonstrate the results of MPS-Net~\cite{wei2022capturing}, GLAMR~\cite{yuan2022glamr} and our method across multiple frames of a video sequence where the person is trying to lift the bag. It can be observed that the head orientation and the proximity of the hand to the bag are more temporally consistent in our method compared to other methods. Further, the SMPL fitting is also better for our method.

\subsection{Generalization to Unseen Datasets}
\label{sec:generaization}
We also test the generalization ability of our method by evaluating its performance on completely unseen Fitness-AQA~\cite{parmar2022domain}, PROX~\cite{hassan2019resolving} and i3DB~\cite{monszpart2019imapper} datasets. These datasets contain diverse scenarios and were not seen during training done on Human3.6M, 3DPW, and MPI-INF-3DHP. As reported in~\autoref{tab:generalization}, our method significantly outperforms existing SOTA on unseen datasets, demonstrating our method's generalization ability. We also show a qualitative comparison for the same in~\autoref{fig:qltv03_ood}. It can be seen that our method estimates pose and shape more accurately than other SOTA methods. 

\subsection{Ablation Study}
\label{sec:ablation}
We perform a detailed ablative study to analyze the contributions of different components of our method.~\autoref{tab:ablation_new} provides the quantitative ablative results, which list our final method's performance in row-6. We sequentially removed each component of our method and reported the performance drop in rows 1-5. More specifically, row-1 reports the results where we replace our body-aware feature encoder with generic ResNet. This leads to a drop in performance, demonstrating the contribution of the body aware features to our overall performance. In row-2, we train our network without using the per-frame pose and camera initialization. This too leads to a drop in the model performance. In row-3 \& row-4, we report the performance of the model by individually removing the pose initialization and camera initialization. The results demonstrate that both pose initialization and camera initialization contribute individually to our method's performance. In row-5, we report the performance by removing the LSTM-based motion refinement component, and once again find a drop in performance, especially in the ACC-ERR metric, demonstrating the contribution of the motion refinement module. We also report two additional ablative results in the last two rows of ~\autoref{tab:ablation_new} as modifications to our proposed method. Specifically, row-7 reports the performance of the modified method by adding the self-attention on the body pose initialization to our method. However, unlike self-attention on body-aware features and camera pose, we empirically find that self-attention on body pose leads to a degradation in performance. One possible explanation for this degradation is that self-attention to body poses can sometimes be misleading due to the frequently repeating body poses in a temporal window (e.g. walking involves very similar body poses). Nevertheless, we observed that using self-similarity (NSSM) on body pose helps as it exploits the spatio-temporal ordering (see row-3 \& row-6). Finally, in row-8, we report the performance of an alternate setup for temporal refinement where we use the LSTM to aggregate temporal features before passing them to the pose/shape predictor, thereby eliminating the coarse prediction step. However, this leads to a drop in performance. As an explanation to this, we hypothesize that learning pose/shape corrections is more conducive to LSTM and hence our method provides a better estimate of body pose.

In addition, we evaluate the performance of our method with different per-frame initialization methods and report results in~\autoref{tab:ablation_diff_init}. It can be observed that our method consistently improves over the per-frame initialization methods (especially in terms of acceleration errors).

Furthermore we examine the performance of our method by varying the choice of temporal window sizes. These results are reported in ~\autoref{tab:ablation_window_size}. Similar to existing works~\cite{kocabas2020vibe, choi2021beyond, wei2022capturing}, we find that a temporal window of size 16 provides optimal performance.

%% file: tables/tab_quant.tex
\begin{table*}[]
    \resizebox{\textwidth}{!}{
        \begin{tabular}{lccc|cccc|ccc}
        \multicolumn{1}{c}{\multirow{2}{*}{Method}} & \multicolumn{3}{c}{Human3.6M~\cite{ionescu2013human3}} & \multicolumn{4}{c}{3DPW~\cite{von2018recovering}} & \multicolumn{3}{c}{MPI-INF-3DHP~\cite{mehta2017monocular}}  \\
        \cline{2-11} 
        \vspace{-0.3cm}
        \\
        
        \multicolumn{1}{c}{} & PA-MPJPE$\downarrow$  & MPJPE $\downarrow$ & ACC-ERR$\downarrow$  & PA-MPJPE$\downarrow$ & MPJPE$\downarrow$ & MPVPE$\downarrow$ & ACC-ERR$\downarrow$ & PA-MPJPE$\downarrow$   & MPJPE$\downarrow$   & ACC-ERR$\downarrow$ \\
        \hline
        VIBE \cite{kocabas2020vibe} & 41.4 & 65.6 & - & 51.9 & 82.9 & 99.1 & 23.4 & 64.6 & 96.6 & - \\
        MEVA  \cite{luo20203d} & 53.2 & 76.0 & 15.3 & 54.7 & 86.9 & - & 11.6 & 65.4 & 96.4 & 11.1 \\
        Uncertainty-Aware \cite{lee2021uncertainty} & 38.4 & 58.4 & 6.1  & 52.2 & 92.8 & 106.1 & 6.8  & \underline{59.4} & \underline{93.5} & 9.4  \\
        TCMR \cite{choi2021beyond} & 52.0 & 73.6 & 3.9 & 52.7 & 86.5 & 102.9 & 7.1 & 63.5 & 97.3 & 8.5 \\
        MPS-Net \cite{wei2022capturing} & 47.4 & 69.4 & \underline{3.6}  & 52.1 & 84.3 & 99.7 & 7.4  & 62.8 & 96.7 & 9.6  \\
        HUMOR \cite{rempe2021humor} & 47.3 & 69.3 & 4.2 & 51.9 & 74.8 & \underline{81.4}  & \underline{6.3} & 63.2 & 98.1 & \underline{8.4}  \\
        GLAMR \cite{yuan2022glamr}$\ast$ & 48.3 & 72.8 & 6.0 & 51.7 & \underline{72.9} & 86.6 & 8.9 & 60.17 & 96.2 & 8.9 \\
        D\&D ~\cite{li2022dnd}$\dagger$ & \underline{35.5} & \underline{52.5} & 6.1 & \underline{42.7} & 73.7 & 88.6 & 7.0 & - & - & - \\
        Our Method & \textbf{31.0} & \textbf{41.3} & \textbf{3.3} & \textbf{39.2} & \textbf{63.5} & \textbf{61.8} & \textbf{5.3} & \textbf{53.2} & \textbf{88.7} & \textbf{8.1} \\ 
        \hline
        \end{tabular}
    }
    \caption{Quantitative Comparison of our method with other monocular video-based methods. Best is in \textbf{bold} and second best is \underline{underlined}. ($\ast$: GLAMR uses Human3.6M, 3DPW and AMASS~\cite{mahmood2019amass} as 3D datasets $|$ $\dagger$: D\&D performs individual training on Human3.6M and 3DPW)}
    \label{tab:quant}
\end{table*}

%% file: tables/tab_generalization.tex
\begin{table}[h]
\centering

\resizebox{0.48\textwidth}{!}{
    \begin{tabular}{lcc|cc|cc}
    \multicolumn{1}{c}{\multirow{2}{*}{Method}} & \multicolumn{2}{c|}{Fitness-AQA~\cite{parmar2022domain}} & \multicolumn{2}{c|}{PROX~\cite{hassan2019resolving}} & \multicolumn{2}{c}{i3dB~\cite{monszpart2019imapper}} \\
    \multicolumn{1}{c}{} & MPJPE $\downarrow$ & ACC-ERR$\downarrow$ & MPJPE$\downarrow$ & ACC-ERR$\downarrow$ & MPJPE$\downarrow$ & ACC-ERR$\downarrow$\\
    \hline
    TCMR~\cite{choi2021beyond} & 89.3 & 7.6 & 29.7 & 2.3 & 47.1 & 3.1\\
    MPS-Net~\cite{wei2022capturing} & 64.7 & 6.1 & 22.1 & 1.9 & 35.5 & 2.7\\ 
    Our Method & \textbf{43.5} & \textbf{5.3} & \textbf{18.3} & \textbf{1.6} & \textbf{21.6} & \textbf{2.1}\\ 
    \hline
    \end{tabular}
}
\caption {Generalization results on unseen datasets.}
\label{tab:generalization}
\end{table}

%% file: tables/tab_ablation_new.tex
\begin{table*}[h]
    \centering
    \resizebox{\textwidth}{!}{
        \begin{tabular}{lccc|ccc|ccc}
        \multirow{2}{*}{Configuration} & \multicolumn{3}{c}{Human3.6M~\cite{ionescu2013human3}} & \multicolumn{3}{c}{3DPW~\cite{von2018recovering}}  & \multicolumn{3}{c}{MPI-INF-3DHP~\cite{mehta2017monocular}} \\
        \cline{2-10} 
        \\
        
         & PA-MPJPE $\downarrow$ & MPJPE $\downarrow$ & ACC-ERR$\downarrow$ & PA-MPJPE $\downarrow$ & MPJPE$\downarrow$ $\downarrow$ & ACC-ERR$\downarrow$ & PA-MPJPE $\downarrow$ &  MPJPE$\downarrow$ & ACC-ERR$\downarrow$   \\
        \hline
        1. Ours \textit{w/o} Body-Aware Features (i.e., \textit{w/o} $H$) & 34.8 & 45.0 & 3.5 & 41.6 & 67.3 & 5.3 & 56.4 & 91.8 & 8.3 \\ 
        2. Ours \textit{w/o} Per-Frame initialization (i.e., \textit{w/o} $\{R^{init}, \theta^{init}\}$ and \textit{w/o} $\omega^{init}$) & 43.8 & 71.9 & 4.2 & 49.8 & 78.5 & 5.5 & 61.4 & 93.4 & 8.4 \\
        3. Ours \textit{w/o} pose initialization (i.e., \textit{w/o} $\{R^{init}, \theta^{init}\}$) & 41.5 & 53.2 & 3.6 & 48.1 & 74.7 & 5.5 & 58.6 & 93.2 & 8.4 \\
        4. Ours \textit{w/o} camera initialization (i.e., \textit{w/o} $\omega^{init}$) & 37.3 & 49.1 & 3.5 & 47.3 & 73.8 & 5.4 & 57.2 & 91.8 & 8.3 \\
        5. Ours \textit{w/o} LSTM based refinement on coarse estimates (i.e., \textit{w/o} $\zeta$) & 32.7 & 42.8 & 3.5 & 40.3 & 69.3 & 5.6 & 54.6 & 90.3 & 8.4 \\
        6. Our Final Method & \textbf{31.0} & \textbf{41.3} & \textbf{3.3} & \textbf{39.2} & \textbf{63.5} & \textbf{5.3} & \textbf{53.2} & \textbf{88.7} & \textbf{8.1}\\ 
        \hdashline
        7. Ours + AM on pose (i.e., AM on $\{R^{init}, \theta^{init}\}$) & 33.8 & 44.8 & 3.4 & 42.7  & 68.0 & 5.3 & 56.8 & 90.7 & 8.3 \\
        8. Ours \textit{w.} LSTM on Feature Space (followed by refined estimation of shape/pose) & 39.2 & 47.2 & 4.0 & 44.3 & 72.9 & 5.8 & 58.7 & 92.6 & 8.3\\
        \hline
        \end{tabular}
    }
    \caption{Ablation study on our method's performance in different configurations. (Best is in \textbf{bold}.)}
    \label{tab:ablation_new}
\end{table*}

%% file: tables/tab_ablation_diff_init.tex
\begin{table*}[]
    \resizebox{\textwidth}{!}{
    \begin{tabular}{lccc|ccc|ccc}
        \multirow{2}{*}{Method} & \multicolumn{3}{c}{Human3.6M~\cite{ionescu2013human3}} & \multicolumn{3}{c}{3DPW~\cite{von2018recovering}} & \multicolumn{3}{c}{MPI-INF-3DHP~\cite{mehta2017monocular}} \\
        \cline{2-10}
        \\
        
         & PA-MPJPE$\downarrow$ & MPJPE$\downarrow$ & ACC-ERR$\downarrow$ & PA-MPJPE$\downarrow$ & MPJPE$\downarrow$ & ACC-ERR$\downarrow$ & PA-MPJPE$\downarrow$ & MPJPE$\downarrow$ & ACC-ERR$\downarrow$ \\ 
        \hline
        PARE~\cite{kocabas2021pare} & 53.8 & 72.8 & 6.9  & 46.5 & 74.5 & 7.1 & 63.1 & 95.7  & 9.3 \\
        Ours w. PARE & 42.6 & 66.4 & 4.1 & 49.3 & 67.3  & 5.4 & 58.7 & 93.3 & 8.1 \\
        CLIFF~\cite{li2022cliff} & 32.7 & 47.1 & 6.7 & 43.0 & 69.0 & 7.3 & 62.4 & 96.7 & 9.3 \\
        Ours w. CLIFF & 31.2 & 42.7 & 5.7 & 39.3 & 65.1 & 6.7 & 59.3 & 94.2 & 8.3 \\
        HMR 2.0~\cite{goel2023humans} & 33.8 & 45.3 & 3.8 & 44.4 & 69.8 & 5.6 & 57.1 & 91.9 & 8.4 \\ 
        Our Method (uses HMR2.0 initialization) & \textbf{31.0} & \textbf{41.3} & \textbf{3.3} & \textbf{39.2} & \textbf{63.5} & \textbf{5.3} & \textbf{53.2} & \textbf{88.7} & \textbf{8.1} \\ 
        \hline 
        \end{tabular}
    }
    \caption{Evaluation of our method with different per-frame initializers. (Best is in \textbf{bold}.)}   \label{tab:ablation_diff_init}
\end{table*}

%% file: tables/tab_ablation_window_size.tex

\begin{table*}[h]
\scriptsize
 \setlength{\tabcolsep}{1.5pt}
\centering
\begin{tabular}{lccc|cccc|ccc}
\multicolumn{1}{c}{\multirow{2}{*}{Window size}} & \multicolumn{3}{c|}{Human3.6M} & \multicolumn{4}{c|}{3DPW} & \multicolumn{3}{c}{MPI-INF-3DHP}  \\
\cline{2-11} \\

\multicolumn{1}{c}{} & PA-MPJPE $\downarrow$ & MPJPE $\downarrow$ & ACC-ERR $\downarrow$ & PA-MPJPE $\downarrow$ & MPJPE $\downarrow$ & MPVPE $\downarrow$ & ACC-ERR $\downarrow$ & PA-MPJPE $\downarrow$ & MPJPE $\downarrow$ & ACC-ERR  \\
    \hline
    8 frames & 31.3 & 41.7 & 3.4 & 40.1 & 64.4 & 62.3 & 5.4 & 54.7 & 89.0 & 8.2 \\
    \hline
    16 frames & \textbf{31.0} & \textbf{41.3} & \textbf{3.3} & \textbf{39.2} & \textbf{63.5} & \textbf{61.8} & \textbf{5.3} & \textbf{53.2} & \textbf{88.7} & \textbf{8.1} \\
    \hline
    32 frames & 32.0 & 42.0 & 3.8 & 40.7 & 65.8 & 63.1 & 5.4 & 55.3 & 89.2 & 8.2 \\
    \hline
    \end{tabular}
    \caption{Ablation study on performance of our method with different temporal window sizes. (Best is in \textbf{bold}).}
\label{tab:ablation_window_size}
\end{table*}

%% file: sections/05_discussion.tex
\begin{figure*}
    \centering
    \includegraphics[width=\linewidth]{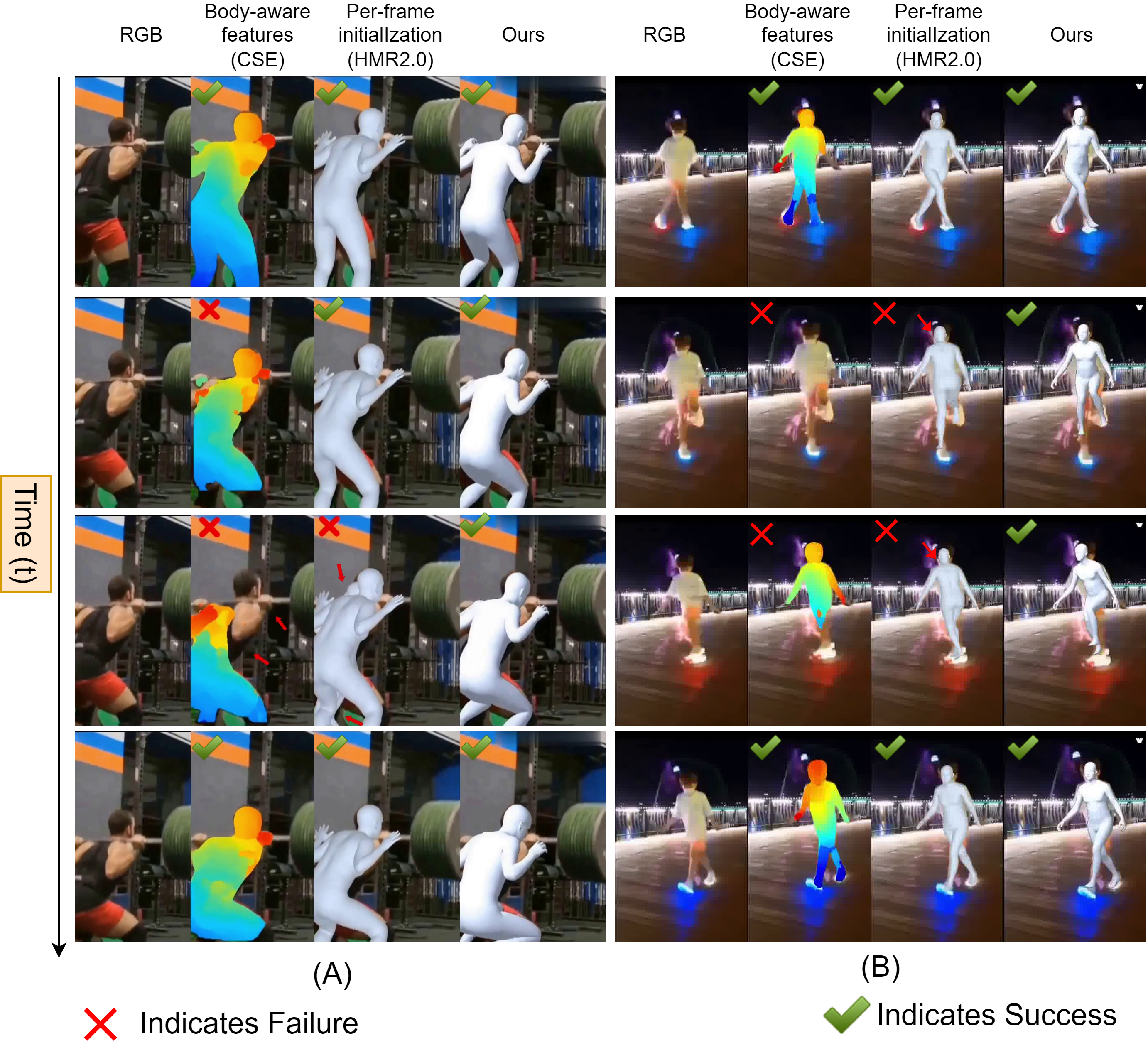}
    \caption{Temporal context provided by STA module allows our method to recover accurate pose and shape even when CSE/HMR2.0 are unable to provide a good initialization. Notice that in the $2^{nd}$ and ${3^{rd}}$ rows of (B), the HMR2.0 prediction is oriented wrongly.}
    \label{fig:temporal_recovery}
\end{figure*}

\begin{figure*}
    \centering
    \includegraphics[width=0.8\textwidth]{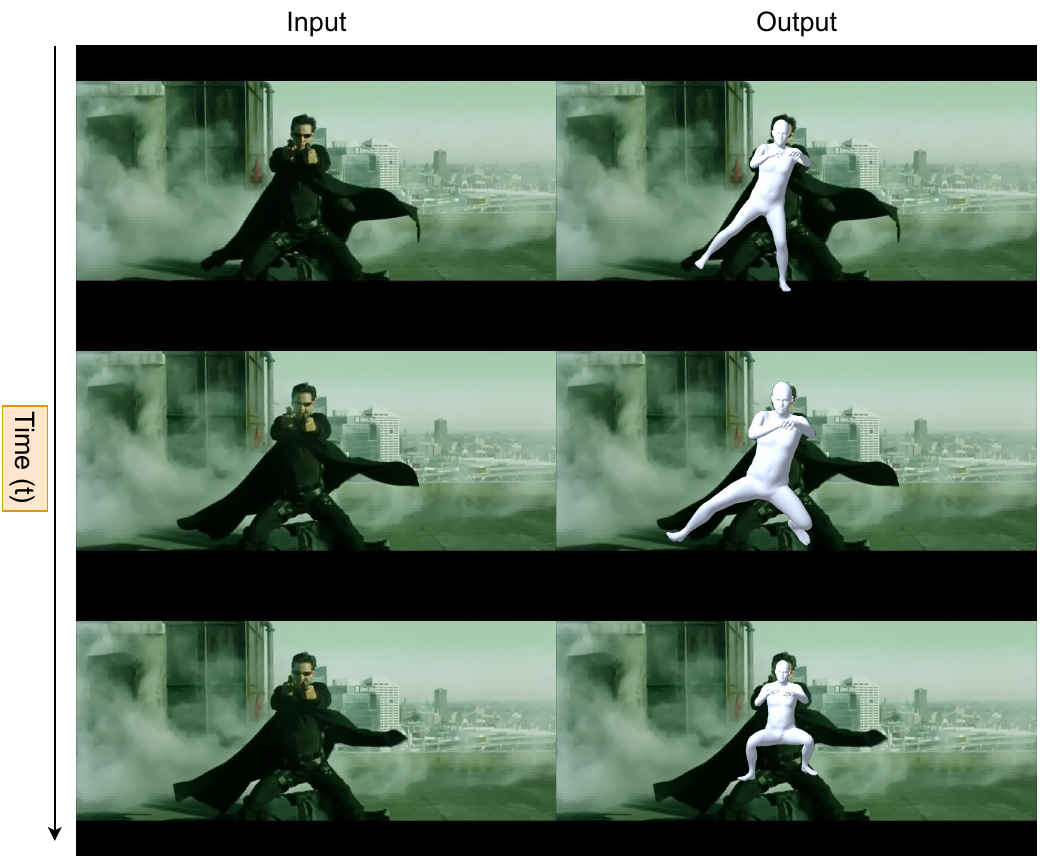}
    \caption{Our method can fail for humans in extremely loose clothing as it is difficult to localize the underlying body in such scenarios.}
    \label{fig:failure_case}
\end{figure*}

\section{Discussion}
\noindent \textbf{Recovering from Bad Initialization:} The spatio-temporal feature aggregation (STA) provides our method temporal context by considering the remaining past and future frames. This allows our method to recover accurate pose and shape even when CSE~\cite{neverova2020continuous} and HMR2~\cite{goel2023humans} are not able to provide good initializations. We show few such results in ~\autoref{fig:temporal_recovery}.

\noindent \textbf{Limitations and Future Work:} As shown in~\autoref{fig:failure_case}, our method can fail in scenarios containing humans with extremely loose clothing as it is difficult to localize the underlying body in such scenarios. We plan to explore extension of our work to loose clothing in the future.

%% file: sections/06_conclusion.tex
\section{Conclusion}
We proposed a novel method for recovering temporally consistent 3D human pose and shape from monocular video. Our method utilizes body-aware spatial features along with initial per-frame SMPL pose parameters to learn spatio-temporally aggregated features over a window. These features are then used to predict the coarse SMPL and camera parameters which are then further refined using a joint prediction of motion with LSTM. We demonstrate that our method consistently outperforms the SOTA methods both qualitatively and quantitatively. We also reported detailed ablative studies to establish relevance of key components of proposed method. As part of future work, it will be interesting to see extension of this work for humans with very loose garments (e.g., robes/abaya).